\documentclass[10pt,twocolumn,letterpaper]{article}

\usepackage{cvpr}
\usepackage{times}
\usepackage{graphicx}
\usepackage{amsmath}
\usepackage{amssymb}
\usepackage{enumerate}
\usepackage{multirow}
\usepackage{boldline}
\usepackage{adjustbox}
\usepackage{subfigure}

\usepackage[pagebackref=true,breaklinks=true,letterpaper=true,colorlinks,bookmarks=false]{hyperref}

\cvprfinalcopy 

\def\cvprPaperID{4763} 

\ifcvprfinal\fi
\begin{document}

\title{Generalized Product Quantization Network for Semi-supervised Image Retrieval}

\author{Young Kyun Jang \and Nam Ik Cho \and\\
Department of ECE, INMC, Seoul National University, Seoul Korea\\
{\tt\small kyun0914@ispl.snu.ac.kr, nicho@snu.ac.kr}
}

\maketitle

\begin{abstract}

Image retrieval methods that employ hashing or vector quantization have achieved great success by taking advantage of deep learning. However, these approaches do not meet expectations unless expensive label information is sufficient. To resolve this issue, we propose the first quantization-based semi-supervised image retrieval scheme: \textbf{G}eneralized \textbf{P}roduct \textbf{Q}uantization (\textbf{GPQ}) network. We design a novel metric learning strategy that preserves semantic similarity between labeled data, and employ entropy regularization term to fully exploit inherent potentials of unlabeled data. Our solution increases the generalization capacity of the quantization network, which allows overcoming previous limitations in the retrieval community. Extensive experimental results demonstrate that GPQ yields state-of-the-art performance on large-scale real image benchmark datasets.
\end{abstract}

\section{Introduction}

The amount of multimedia data, including images and videos, increases exponentially on a daily basis. Hence, retrieving relevant content from a large-scale database has become a more complicated problem. There have been many kinds of fast and accurate search algorithms, and the Approximate Nearest Neighbor (ANN) search is known to have high retrieval accuracy and computational efficiency. Recent ANN methods mainly focused on \textit{hashing} scheme ~\cite{Survey}, because of its low storage cost and fast retrieval speed. To be specific, an image is represented by a binary-valued compact hash code (\textit{binary code}) with only a few tens of bits, and it is utilized to build database and distance computation.

\begin{figure}[!t]
\centering
\includegraphics[width=0.95\linewidth]{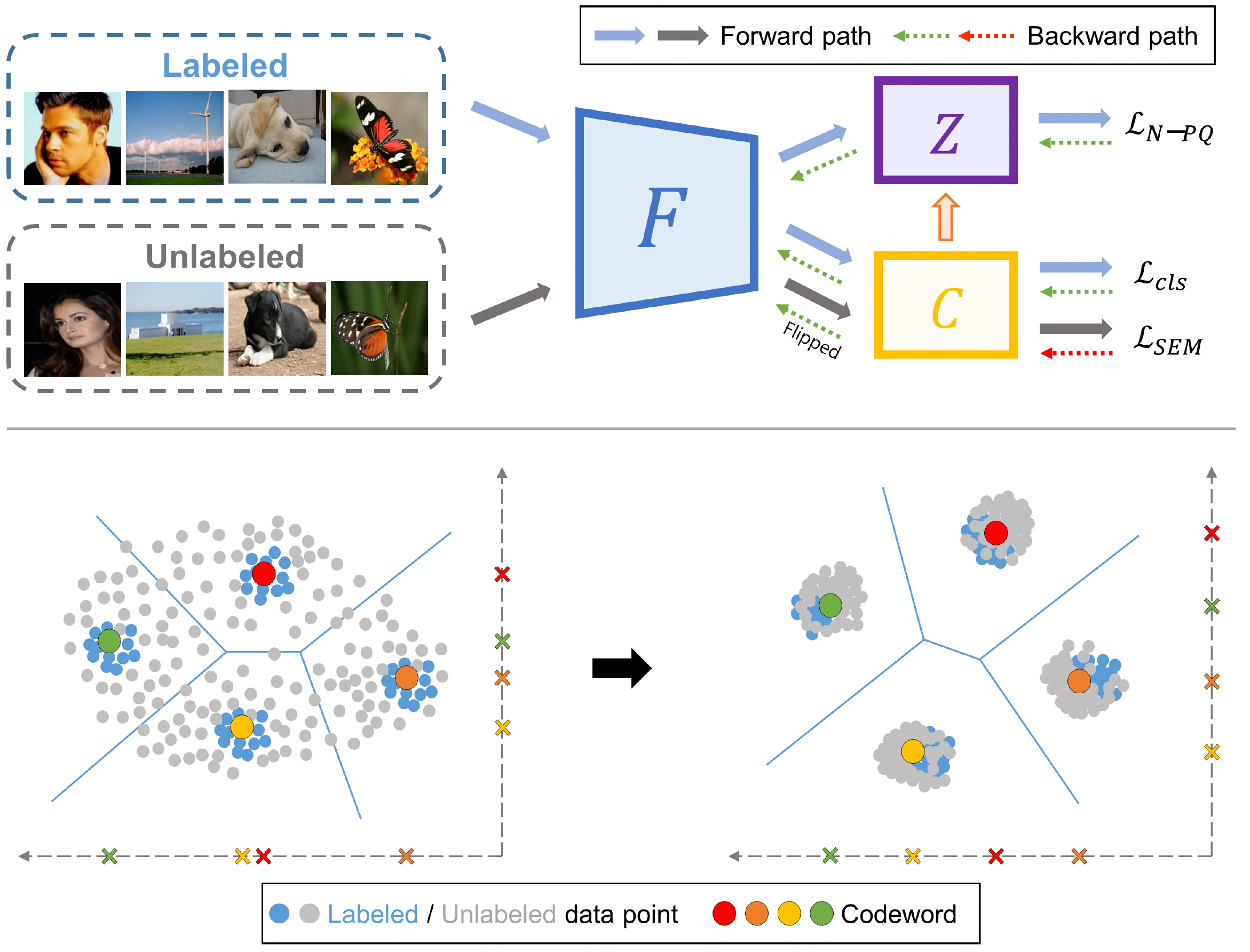}
\caption{Above: an illustration of the overall framework of GPQ and its three components: feature extractor $F$, PQ table $Z$, and classifier $C$, where $C$ contributes to build $Z$. Forward path shows how the labeled and unlabeled data pass through the network, and backward path shows the propagation of the gradients originated from the training objectives. $\mathcal{L}_{N\text{-}PQ}$ and $\mathcal{L}_{cls}$ train the network to minimize errors using the labeled data, while $\mathcal{L}_{SEM}$ trains the network to simultaneously maximize and minimize entropy using the unlabeled data. Below: a 2D conceptual Voronoi diagram showing one of the codebooks in GPQ. After training, all codewords are evenly distributed, and both labeled and unlabeled data points are clustering around them.} 
\label{fig:long}
\label{fig:onecol}
\label{fig:Figure1}
\end{figure}

The methods using binary code representation can be categorized as \textit{Binary hashing} (BH) and \textit{Product Quantization} (PQ) ~\cite{PQ}. BH-based methods~\cite{SH, ITQ, SDH} employ a hash function that maps a high-dimensional vector space to a Hamming space, where the distance between two codes can be measured extremely fast via bitwise XOR operation. However, BH has a limitation in describing the distance between data points because it can produce only a limited number of distinct values. PQ, which is a kind of vector quantization~\cite{VQ}, has been introduced to alleviate this problem in information retrieval ~\cite{PQ, OPQ, LOPQ}.

To perform PQ, we first need to decompose the input feature space into a Cartesian product of several disjoint subspaces (\textit{codebooks}) and find the centroid (\textit{codeword}) of each subspace. Then, from the sub-vectors of the input feature vector, sub-binary code is obtained by replacing each sub-vector with the index of the nearest codeword in the codebook. Since codeword consists of real numbers, PQ allows asymmetric distance calculation in real space using the binary codes, making many PQ-based approaches outperform BH-based ones.

Along with millions of elaborately labeled data, \textit{deep hashing} for both BH ~\cite{CNNH, NINH, SUBIC, DCBH, SHAN} and PQ ~\cite{DQN, DTQ, PQN, DPQ} has been introduced to take advantage of deep representations for image retrieval. By employing supervised deep neural networks, deep hashing outperforms conventional ones on many benchmark datasets. Nevertheless, there still is a great deal of potential for improvement, since a significant amount of unlabeled data with abundant knowledge is not utilized. To resolve these issue, some recent methods are considering the \textit{Deep Semi-Supervised Hashing}, based on BH ~\cite{SSDH, BGDH, SSGAH}. However, even if PQ generally outperforms BH for both supervised and unsupervised settings, it has not yet been considered for learning in a semi-supervised manner. In this paper, we propose the first PQ-based deep semi-supervised image retrieval approach: Generalized Product Quantization (GPQ) network, which significantly improves the retrieval accuracy with lots of image data and just a few labels per category (class).

Existing deep semi-supervised BH methods construct graphs ~\cite{SSDH,BGDH} or apply additional generative models ~\cite{SSGAH} to encode unlabeled data into the binary code. However, due to the fundamental problem in BH; a deviation that occurs when embedding a continuous deep representation into a discrete binary code restricts extensive information of unlabeled data. In our GPQ framework, this problem is solved by including quantization process into the network learning. We adopt intra-normalization~\cite{Intra-Norm} and soft assignment ~\cite{PQN} to quantize real-valued input sub-vectors, and introduce an efficient metric learning strategy; ~\textit{N-pair Product Quantization loss} inspired from ~\cite{N-pair}. By this, we can embed multiple pair-wise semantic similarity between every feature vectors in a training batch into the codewords. It also has an advantage of not requiring any complex batch configuration strategy to learn pairwise relations.

The key point of deep semi-supervised retrieval is to avoid overfitting to labeled data and increase the generalization toward unlabeled one. For this, we suggest a ~\textit{Subspace Entropy Mini-max Loss} for every codebook in GPQ, which regularizes the network using unlabeled data. Precisely, we first learn a cosine similarity-based classifier, which is commonly used in few-shot learning~\cite{Fewshot1, Fewshot2}. The classifier has as many weight matrices as the number of codebooks, and each matrix contains class-specific weight vector, which can be regarded as a \textit{sub-prototype} that indicates class representative centroid of each codebook. Then, we compute the entropy between the distributions of sub-prototypes and unlabeled sub-vectors by measuring their cosine similarity. By maximizing the entropy, the two distributions become similar, allowing the sub-prototypes to move closer to the unlabeled sub-vectors. At the same time, we also minimize the entropy of the distribution of the unlabeled sub-vectors, making them assemble near the moved sub-prototypes. With the gradient reversal layer generally used for deep domain adaptation ~\cite{DA1, DA2}, we are able to simultaneously minimize and maximize the entropy during the network training.

In summary, the main contributions of our work are as follows:
\begin{itemize}
\item To the best of our knowledge, our work is the first deep semi-supervised PQ scheme for image retrieval.
\item With the proposed metric learning strategy and entropy regularization term, the semantic similarity of labeled data is well preserved into the codewords, and the underlying structure of unlabeled data can be fully used to generalize the network.
\item Extensive experimental results demonstrate that our GPQ can yield the state-of-the-art retrieval results in semi-supervised image retrieval protocols.
\end{itemize}

\section{Related Work}

\noindent\textbf{Existing Hashing Methods} Referring to the survey \cite{Survey}, early works in Binary Hashing (BH) ~\cite{SH, ITQ, SDH, SPLH} and Product Quantization (PQ) ~\cite{PQ, OPQ, LOPQ, DEPQ, SSPQ, BOPQ, QVR} mainly focused on unsupervised settings. Specifically, Spectral Hashing (SH)~\cite{SH} considered correlations within hash functions to obtain balanced compact codes. Iterative Quantization (ITQ)~\cite{ITQ} addressed the problem of preserving the similarity of original data by minimizing the quantization error in hash functions. There were several studies to improve PQ, for example, Optimized Product Quantization (OPQ)~\cite{OPQ} tried to improve the space decomposition and codebook learning procedure to reduce the quantization error. Locally Optimized Product Quantization (LOPQ) \cite{LOPQ} employed a coarse quantizer with locally optimized PQ to explore more possible centroids. These methods might reveal some distinguishable results, however, they still have disadvantage of not exploiting expensive label signals.

\bigskip

\noindent\textbf{Deep Hashing Methods} After Supervised Discrete Hashing (SDH) \cite{SDH} has shown the capability to improve using labels, supervised Convolutional Neural Network (CNN)-based BH approaches ~\cite{CNNH, NINH, SUBIC, DCBH, SHAN}are leading the mainstream. For examples, CNN Hashing (CNNH)~\cite{CNNH} utilized a CNN to simultaneously learn feature representation and hash functions with the given pairwise similarity matrix. Network in Network Hashing (NINH)~\cite{NINH} introduced a sub-network, divide-and-encode module, and a triplet ranking loss for the similarity-preserving hashing. A Supervised, Structured Binary Code (SUBIC)~\cite{SUBIC}, used a block-softmax nonlinear function and computed batch-based entropy error to embed the structure into a binary form. There also have been researches on supervised learning that uses PQ with CNN ~\cite{DQN, DTQ, PQN, DPQ}. Precisely, Deep Quantization Network (DQN)~\cite{DQN} simultaneously optimizes a pairwise cosine loss on semantic similarity pairs to learn feature representations and a product quantization loss to learn the codebooks. Deep Triplet Quantization (DTQ)~\cite{DTQ} designed a group hard triplet selection strategy and trained triplets by triplet quantization loss with weak orthogonality constraint. Product Quantization Network (PQN)~\cite{PQN} applied the asymmetric distance calculation mechanism to the triplets and exploited softmax function to build a differentiable soft product quantization layer to train the network in an end-to-end manner. Our method is also based on PQ, but we try a {\em semi}-supervised PQ scheme that had not been considered previously.

\bigskip

\noindent\textbf{Deep Semi-supervised Image Retrieval} Assigning labels to images is not only expensive but also has the disadvantage of restricting the data structure to the labels. Deep semi-supervised hashing based on BH is being considered in the image retrieval community to alleviate this problem, with the use of a small amount of labeled data and a large amount of unlabeled data. For example, Semi-supervised Deep Hashing (SSDH)~\cite{SSDH} employs an online graph construction strategy to train a network using unlabeled data. Deep Hashing with a Bipartite Graph (BGDH)~\cite{BGDH} improved SSDH by using the bipartite graph, which is more efficient in building a graph and learning the embeddings. Since Generative Adversarial Network (GAN) had been used for BH and showed good performance as in~\cite{SHAN}, Semi-supervised Generative Adversarial Hashing (SSGAH) also employed the GAN to fully utilize triplet-wise information of both labeled and unlabeled data. In this paper, we propose GPQ, the first deep semi-supervised image retrieval method applying PQ. In our work, we endeavor to generalize the whole network by preserving the semantic similarity with N-pair Product Quantization loss and extracting the underlying structure of unlabeled data with the Subspace Entropy Mini-max loss.

\begin{figure*}[!t]
\centering
\includegraphics[width=0.85\linewidth]{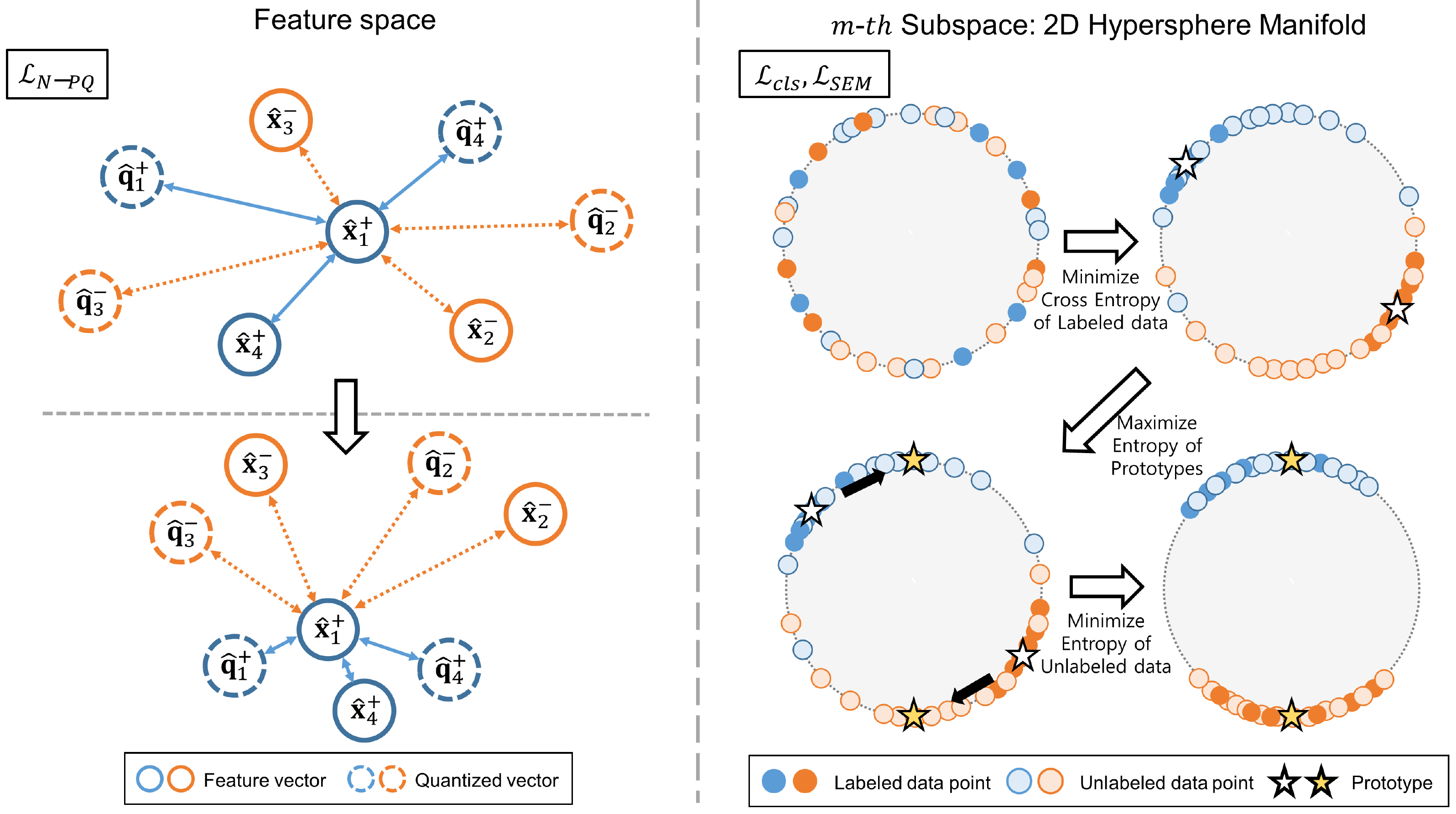}
\caption{A two class ($+$: blue, $-$: orange) visualized examples of our training objectives. 1) The left part shows the learning process of N-pair Product Quantization loss $\mathcal{L}_{N\text{-}PQ}$. When we define an anchor as $\hat{\mathbf{x}}^{+}_1$, the semantically similar points $(\hat{\mathbf{q}}^{+}_1, \hat{\mathbf{x}}^{+}_4, \hat{\mathbf{q}}^{+}_4)$ are pulled together while the semantically dissimilar points $(\hat{\mathbf{x}}^{-}_2, \hat{\mathbf{q}}^{-}_2, \hat{\mathbf{x}}^{-}_3, \hat{\mathbf{q}}^{-}_3)$ are pushing the anchor. 2) The right part shows the learning process of classification loss $\mathcal{L}_{cls}$ and subspace entropy mini-max loss $\mathcal{L}_{SEM}$. For the data points constrained on the unit hypersphere, the cross entropy of labeled data points is minimized to find prototypes (white stars). Then, the entropy between the prototypes and the unlabeled data points is maximized to move prototypes toward unlabeled data points and find new prototypes (yellow stars). Finally, the entropy of the unlabeled data points is minimized to cluster them near the new prototypes.} 
\label{fig:long}
\label{fig:onecol}
\label{fig:Figure2}
\end{figure*}

\section{Generalized Product Quantization}

Given a dataset $\mathcal{X}$ that is composed of individual images, we split this into two subsets as a labeled dataset $\mathcal{X}^L=\{(I^L_i, y_i)|i=1,...,N^L\}$ and an unlabeled dataset $\mathcal{X}^U=\{I^U_i|i=1,...,N^U\}$ to establish a semi-supervised environment. The goal of our work is learning a quantization function $\mathrm{q}:I\rightarrow{\hat{\mathbf{b}}}\in\{0,1\}^{\mathcal{B}}$ which maps a high-dimensional input $I$ to a compact $\mathcal{B}$-bits binary code $\hat{\mathbf{b}}$, by utilizing both labeled and unlabeled datasets. We propose a semi-supervised deep hashing framework: GPQ, which integrates $\mathrm{q}$ into the deep network as a form of Product Quantization (PQ) ~\cite{PQ} to learn the deep representations and the codewords jointly. In the learning process, we aim to preserve the semantic similarity of labeled data and simultaneously explore the structures of unlabeled data to obtain high retrieval accuracy.

GPQ contains three trainable components: 1) a standard deep convolutional neural network-based feature extractor $F$, e.g. AlexNet ~\cite{AlexNet}, CNN-F ~\cite{CNN-F} or modified version of VGG \cite{DCBH} to learn deep representations; 2) a PQ table $Z$ that collects codebooks which are used to map an extracted feature vector to a binary code; 3) a cosine similarity-based classifier $C$ to classify both labeled and unlabeled data. GPQ network is designed to train all these components in an end-to-end manner. In this section, we will describe each component and how GPQ is learned in a semi-supervised way.

\subsection{Semi-Supervised Learning}

The feature extractor $F$ generates $D$-dimensional feature vector $\hat{\mathbf{x}} \in R^D$. Under semi-supervised learning condition, we aim to train the $F$ to extract discriminative $\hat{\mathbf{x}}^L$ and $\hat{\mathbf{x}}^U$ from labeled image $I^L$ and unlabeled image $I^U$, respectively. Also, we leverage the PQ concept to utilize these feature vectors for image retrieval, which requires appropriate codebooks with distinct codewords to replace and store the feature vectors. We introduce three training objectives for our GPQ approach to fully exploit the data structure of labeled and unlabeled images, and we illustrate a conceptual visualization of each loss function in Figure \ref{fig:Figure2} for better understanding.

Following the observation of~\cite{Fewshot1,Fewshot2, PQN, DTQ}, we normalize the feature vectors and constrain them on a unit hypersphere to focus on the angle rather than the magnitude in measuring the distance between two different vectors. In this way, every data is mapped to the nearest class representative direction, and better performs for the semi-supervised scheme because the distribution divergence between labeled and unlabeled data can be reduced within the constraint.
Especially for PQ, we apply intra-normalization \cite{Intra-Norm} for a feature vector $\hat{\mathbf{x}}$ by dividing it into $M$-sub-vectors $\hat{\mathbf{x}}=[\mathbf{x}_1,...,\mathbf{x}_M],$ where $\mathbf{x}_m \in R^{d}, d=D/M$, and ${l2}$-normalize each sub-vector as: $\mathbf{x}_m\leftarrow{\mathbf{x}_m/||\mathbf{x}_m||_2}$. In the rest of the paper, $\hat{\mathbf{x}}$ for GPQ denotes the intra-normalized feature vector.

\bigskip

\noindent\textbf{N-pair Product Quantization} The PQ table $Z$ collects $M$-codebooks $Z=[\mathbf{Z}_1,...,\mathbf{Z}_M]$ and each codebook has $K$-codewords ~$\mathbf{Z}_m=[\mathbf{z}_{m1},...,\mathbf{z}_{mK}]$ where $\mathbf{z}_{mk} \in R^d$, which are used to replace $\hat{\mathbf{x}}$ with the quantized vector $\hat{\mathbf{q}}$. Every codeword is ${l2}$-normalized to simplify the measurement of cosine similarity as multiplication. We employ the soft assignment $s_m(\cdot)$ ~\cite{PQN} to obtain a $\mathbf{q}_m$ from $\mathbf{x}_m$ as:

\begin{align}
\mathbf{q}_m=\sum_{k}^{K}\frac{e^{-\alpha(\mathbf{x}_m\cdot \mathbf{z}_{mk})}}{\sum_{k'}^{K}e^{-\alpha(\mathbf{x}_m\cdot \mathbf{z}_{mk'})}}\mathbf{z}_{mk}
\label{equation:Eqn1}
\end{align}

\noindent where $\alpha$ represents a scaling factor to approximate hard assignment, and ~$\mathbf{q}_m=s_m(\mathbf{x}_m;\alpha, \mathbf{Z}_m)$ is the sub-quantized vector of ~$\hat{\mathbf{q}}=[\mathbf{q}_1,...,\mathbf{q}_M]$. We multiply $\mathbf{x}_m$ with the every codeword in $\mathbf{Z}_m$ to measure the cosine similarity between them.

Quantization error occurs through the encoding process; therefore, we need to find the codewords that will minimize the error. In addition, the conventional PQ scheme has a limitation that it ignores label information since the sub-vectors are clustered to find codewords without any supervised signals. To fully exploit the semantic labels and reduce the quantization error, we revise the metric learning strategy proposed in \cite{N-pair} from \textit{N-pair Product Quantization loss}: $\mathcal{L}_{N\text{-}PQ}$ to learn the $F$ and $Z$, and set it as one of the training objectives.

Deep metric learning \cite{Metric1, Metric2} aims to learn an embedding representation of the data with the semantic labels. From a labeled image $I^L$, we can generate a unique feature vector $\hat{\mathbf{x}}^L$ and its nearest quantized vector $\hat{\mathbf{q}}^L$, which can be regarded as sharing the same semantic information. Thus, for randomly sampled $B$ training examples $\{(I^L_1,y_1),...,(I^L_B,y_B)\}$, the objective function which is based on a standard cross entropy loss $\mathcal{L}_{CE}$, can be formulated as:

\begin{align}
\mathcal{L}_{N\text{-}PQ}=\frac{1}{B}\sum_{b=1}^{B}\mathcal{L}_{CE}(\mathcal{S}_b, \mathbf{Y}_b)
\label{equation:Eqn2}
\end{align}

\noindent where $\mathcal{S}_b=\left[(\hat{\mathbf{x}}^L_b)^T \hat{\mathbf{q}}^L_1,...,(\hat{\mathbf{x}}^L_b)^T \hat{\mathbf{q}}^L_B\right]$ denotes a cosine similarity between $b$-th feature vector and every quantized vector, and $\mathbf{Y}_b=\left[(\mathbf{y}_b)^T \mathbf{y}_1,...,(\mathbf{y}_b)^T \mathbf{y}_B\right]$ denotes a similarity between $b$-th label and every label in a batch. In this case, $\mathbf{y}$ represents one-hot-encoded semantic label, and column-wise normalization is applied to $\mathbf{Y}_B$. $\mathcal{L}_{N\text{-}PQ}$ has an advantage in that no complicated batch construction method is required, and it also allows us to jointly learn the deep feature representation for both feature vectors and the codewords on the same embedding space.

\bigskip

\noindent\textbf{Cosine Similarity-based Classification} To embed semantic information into the codewords while reducing correlation between each codebook, we learn a cosine similarity-based classifier $C$ containing $M$-weight matrices $[\mathbf{W}_1,...,\mathbf{W}_M]$, where each matrix includes sub-prototypes as ~$\mathbf{W}_m=[\mathbf{c}_{m1},...,\mathbf{c}_{mN^c}]$, $\mathbf{W}_m\in R^{d\times N^c}$ and $N^c$ is the number of class. Every sub-prototype is ${l2}$-normalized to hold a class-specific angular information. With the $m$-th sub-vector $\mathbf{x}_m^L$ of $\mathbf{x}^L$ and $m$-th weight matrix $\mathbf{W}_m$, we can obtain the labeled class prediction as: $\mathbf{p}_m^L=\mathbf{W}_m^T\mathbf{x}_m^L$. We use $\mathcal{L}_{CE}$ again to train the $F$ and $C$ for classification using $\hat{\mathbf{p}}^L=[\mathbf{p}^L_1,...,\mathbf{p}^L_M]$ computed from $x^L$ with the corresponding semantic label $y$ as:

\begin{align}
\mathcal{L}_{cls}=\frac{1}{M}\sum_{m=1}^{M}\mathcal{L}_{CE}(\beta \cdot \mathbf{p}_m^L, y)
\label{equation:Eqn3}
\end{align}

\noindent where $\beta$ is a scaling factor and $y$ is a label corresponding to $x^L$. This classification loss ensures the feature extractor to generate the discriminative features with respect to the labeled examples. In addition, each weight matrix in the classifier is derived to include the class-specific representative sub-prototypes of related subspace.

\bigskip

\noindent\textbf{Subspace Entropy Mini-max} On the assumption that the distribution is not severely different between the labeled data and the unlabeled data, we aim to propagate the gradients derived from the divergence between them. To calculate the error arising from the distribution difference, we adopt an entropy in information theory. In particular for PQ setting, we compute the entropy for each subspace to balance the amount of gradients propagating into each subspace. With the $m$-th sub-vector $\mathbf{x}_m^U$ of the unlabeled feature vector $\mathbf{x}^U$, and the $m$-th weight matrix of $C$ that embraces sub-prototypes, we can obtain a class prediction using cosine similarity as: $\mathbf{p}^U_m=\mathbf{W}_m^T\mathbf{x}_m^U$. By using it, the subspace entropy mini-max loss is calculated as:

\begin{align}
\mathcal{L}_{SEM}=-\frac{1}{M}\sum_{m=1}^{M}\sum_{l=1}^{N^c} (\beta \cdot p_{ml}^{U})\log(\beta \cdot p_{ml}^{U})
\label{equation:Eqn4}
\end{align}

\noindent where $\beta$ is the same as that in Equation ~\ref{equation:Eqn3} and $p_{ml}^{U}$ denotes the probability of prediction to $k'$-th class; $l$-th element of $\mathbf{p}^U_m$. The generalization capacity of the network can increase by maximizing the $\mathcal{L}_{SEM}$ because high entropy ensures that the sub-prototypes are regularized toward unlabeled data. Explicitly, entropy maximization makes sub-prototypes have a similar distribution with the unlabeled sub-vectors, moving sub-prototypes near the unlabeled sub-vectors. To further improve, we aim to cluster unlabeled sub-vectors near the moved sub-prototype, by applying gradient reversal layer~\cite{DA1, DA2} before intra-normalization. Flipped gradients induce $F$ to be learned in the direction of minimizing the entropy, resulting in a skewed distribution of unlabeled data.

From the Equations~\ref{equation:Eqn2} to~\ref{equation:Eqn4}, total objective function $\mathcal{L}_{T}$ for $B$ randomly sampled training pairs of $I^L$ and $I^U$, can be formulated as:

\begin{align}
\mathcal{L}_{T}(\mathbf{B})=\mathcal{L}_{N\text{-}PQ}&+\frac{1}{B}\sum_{b=1}^{B}\left(\lambda_{1}\mathcal{L}_{cls}-\lambda_{2}\mathcal{L}_{SEM})\right.
\label{equation:Eqn5}
\end{align}

\noindent where $\mathbf{B} = \{(I^L_1,y_1,I^U_1),...,(I^L_B,y_B,I^U_B)\}$ and $\lambda_1$ and $\lambda_2$ are the hyper-parameters that balance the contribution of each loss function. We force training optimizer to minimize the $\mathcal{L}_{T}$, so that the $\mathcal{L}_{N-PQ}$ and $\mathcal{L}_{cls}$ is minimized while the $\mathcal{L}_{SEM}$ is maximized, simultaneously. In this way, $F$ can learn the deep representation of both labeled and unlabeled data. However, to make the codewords robust against unlabeled data, it is necessary to reflect the unlabeled data signals directly into $Z$. Accordingly, we apply another soft assignment to embed sub-prototype intelligence of $\mathbf{W}_m$ into the of $m$-th codebook by updating codewords as $\mathbf{z}'_{mk}=s_m\left(\mathbf{z}_{mk};\alpha, \mathbf{W}_m\right)$. As a result, high retrieval performance can be expected by exploiting the potential of unlabeled data for quantization.

\subsection{Retrieval}

\noindent\textbf{Building Retrieval Database} After learning the entire GPQ framework, we can build a retrieval database using images in $\mathcal{X}^U$. Given an input image $I^R\in \mathcal{X}^U$, we first extract $\hat{\mathbf{x}}^R$ from $F$. Then, we find the nearest codeword $\mathbf{z}_{mk^*}$ of each sub-vector $\mathbf{x}_m^R$ from the corresponding codebook $\mathbf{Z}_m$, by computing the cosine similarity. After that, formatting a index $k^*$ of the nearest codeword as binary to generate a sub-binary code $\mathbf{b}^R$. Finally, concatenate all the sub-binary codes to obtain a $M{\cdot}\log_2(K)$-bits binary code $\hat{\mathbf{b}}^R$, where $\hat{\mathbf{b}}^R=[\mathbf{b}_1^R,...,\mathbf{b}_M^R]$. This procedure is repeated for all images to store them as binary, and $Z$ is also stored for distance calculation.

\bigskip

\noindent\textbf{Asymmetric Search} For a given query image $I^Q$, $\hat{\mathbf{x}}^Q$ is extracted from $F$. To conduct the image retrieval, we take the $m$-th sub-vector $\mathbf{x}_m^Q$ as an example, compute the cosine similarity between $\mathbf{x}_m^Q$ and every codeword belonging to the $m$-th codebook, and store measured similarities on the look-up-table (LUT). Similarly, the same operation is done for the other sub-vectors, and the results are also stored on the LUT. The distance between the query image and a binary code in the database can be calculated asymmetrically, by loading the pre-computed distance from the LUT using a sub-binary codes, and aggregating all the loaded distances.

\section{Experiments}

We evaluate GPQ for two semi-supervised image retrieval protocols against several hashing approaches. We conduct experiments on the two most popular image retrieval benchmark datasets. Extensive experimental results show that GPQ achieves superior performance to existing methods.

\subsection{Setup}
\noindent\textbf{Evaluation Protocols and Metrics} Following the semi-supervised retrieval experiments in ~\cite{SSDH, BGDH, SSGAH}, we adopt two protocols as follows.

\begin{enumerate}[-]
\item \textbf{Protocol 1: Single Category Image Retrieval} Assuming all categories (classes) used for image retrieval are known and only a small number of label data is provided for each class. The labeled data is used for training, and the unlabeled data is used for building the retrieval database and the query dataset. In this case, labeled data and the unlabeled data belonging to the retrieval database are used for semi-supervised learning.

\item \textbf{Protocol 2: Unseen Category Image Retrieval} In line with semi-supervised learning, suppose that the information of the categories in the query dataset is unknown, and consider building a retrieval database using both known and unknown categories. For this situation, we divide image dataset into four parts: train75, test75, train25, and test25, where train75 and test75 are the data of 75\% categories, while train25 and test25 are the data of the remaining 25\% categories. We use train75 for training, train25 and test75 for the retrieval database, and test25 for the query dataset. In this case, train75 with labels, and train25, test75 without labels are used for semi-supervised learning.

\end{enumerate}

\noindent The retrieval performance of hashing method is measured by mAP (mean Average Precision) with bit lengths of 12, 24, 32, and 48 for all images in the query dataset. In particular, we set Protocol 1 as the primary experiment and observe the contribution of each training objective.

\bigskip

\noindent\textbf{Datasets} We set up two benchmark datasets, different for each protocol as shown in the Table ~\ref{table:Table1}, and each dataset is configured as follows.

\begin{table}[!b]
\centering
\begin{adjustbox}{width=0.45\textwidth}
\small
\begin{tabular}{|c|c|c|c|c|}
\hline
\multirow{2}{*}{}                                            & \multicolumn{2}{c|}{CIFAR-10}                            & \multicolumn{2}{c|}{NUS-WIDE}                            \\ \cline{2-5} 
                                                             & \multicolumn{1}{l|}{Protocol 1} & \multicolumn{1}{l|}{Protocol 2} & \multicolumn{1}{l|}{Protocol 1} & \multicolumn{1}{l|}{Protocol 2} \\ \hline
Query                                                    & 1,000                           & 9,000                           & 2,100                           & 35,272                          \\ \hline
Training                                                 & 5,000                           & 21,000                          & 10,500                          & 48,956                          \\ \hline
\begin{tabular}[c]{@{}c@{}}Retrieval\\ Database\end{tabular} & 54,000                          & 30,000                          & 157,043                        & 85,415                          \\ \hline
\end{tabular}
\end{adjustbox}
\caption{Detailed composition of two benchmark datasets.}
\label{table:Table1}
\end{table}

\begin{enumerate}[-]
\item \textbf{CIFAR-10} is a dataset containing 60,000 color images with the size of $32\times32$. Each image belongs to one of 10 categories, and each category includes 6,000 images.
\item \textbf{NUS-WIDE} ~\cite{NUS-WIDE} is a dataset consisting nearly 270,000 color images with various resolutions. Images in the dataset associate with one or more class labels of 81 semantic concepts. We select the 21 most frequent concepts for experiments, where each concept has more than 5,000 images, with a total of 169,643.

\end{enumerate}

\noindent\textbf{Implementation Details} \label{Implementation Details} We implement our GPQ based on the Tensorflow framework and perform it with a NVIDIA Titan XP GPU. When it comes to conducting experiments on non-deep learning-based hashing methods ~\cite{SH,ITQ,SDH,PQ,OPQ,LOPQ}, we utilize hand-crafted features as inputs following ~\cite{NINH}, extracting 512-dimensional GIST ~\cite{GIST} features from CIFAR-10 images, and 500-dimensional bag-of-words features from NUS-WIDE images. For deep hashing methods ~\cite{CNNH, NINH, SUBIC, DQN, DTQ, PQN, SSDH, BGDH, SSGAH}, we use raw images as inputs and adopt ImageNet pretrained AlexNet ~\cite{AlexNet} and CNN-F ~\cite{CNN-F} as the backbone architectures to extract the deep representations. Since AlexNet and CNN-F significantly downsample the input image at the first convolution layer, they are inadequate especially for small images. Therefore, we employ modified VGG architecture proposed in ~\cite{DCBH} as the baseline architecture for feature extractor of GPQ. For a fair comparison, we also employ CNN-F to evaluate our scheme (GPQ-F), and details will be discussed in the section ~\ref{Ablation Study}.

With regard to network training, we adopt ADAM algorithm to optimize the network, and apply exponential learning rate decay with the initial value 0.0002 and the ~$\beta_1=0.5$. We configure the number of labeled and unlabeled images equally in the learning batch. To simplify the experiment, we set several hyper-parameters as follows: the scaling factors $\alpha$ and $\beta$ are fixed for 20 and 4 respectively, the codeword number $K$ is fixed at $2^4$ while $M$ is adjusted to handle multiple bit lengths, and the dimension $d$ of the sub-vector $\mathbf{x}_m$ is fixed at 12. Detail analysis of these hyper-parameters is shown in the supplementary. We upload our method in \url{https://github.com/youngkyunJang/GPQ}.

\begin{table*}[!t]
\centering
\begin{adjustbox}{width=0.9\textwidth}
\tiny
\begin{tabular}{|c|c|c|c|c|c|c|c|c|c|}
\hline
\multirow{2}{*}{Concept}              & \multirow{2}{*}{Method} & \multicolumn{4}{c|}{CIFAR-10}                                     & \multicolumn{4}{c|}{NUS-WIDE}                                     \\ \cline{3-10} 
                                      &                         & 12-bits        & 24-bits        & 32-bits        & 48-bits        & 12-bits        & 24-bits        & 32-bits        & 48-bits        \\ \hline
\multirow{4}{*}{Deep Semi-supervised} & GPQ (Ours)                    & \textbf{0.858} & \textbf{0.869} & \textbf{0.878} & \textbf{0.883} & \textbf{0.852} & \textbf{0.865} & \textbf{0.876} & \textbf{0.878} \\ \cline{2-10} 
                                      & SSGAH \cite{SSGAH}                   & 0.819          & 0.837          & 0.847          & 0.855          & 0.838          & 0.849          & 0.863          & 0.867          \\ \cline{2-10} 
                                      & BGDH \cite{BGDH}                    & 0.805          & 0.824          & 0.826          & 0.833          & 0.810          & 0.821          & 0.825          & 0.829          \\ \cline{2-10} 
                                      & SSDH \cite{SSDH}                   & 0.801          & 0.813          & 0.812          & 0.814          & 0.783          & 0.788          & 0.791          & 0.794          \\ \hline
\multirow{3}{*}{Deep Quantization}    & PQN \cite{PQN}                    & 0.795          & 0.819          & 0.823          & 0.830          & 0.803          & 0.818          & 0.822          & 0.824          \\ \cline{2-10} 
                                      & DTQ \cite{DTQ}                    & 0.785          & 0.789          & 0.790          & 0.792          & 0.791          & 0.798          & 0.808          & 0.811          \\ \cline{2-10} 
                                      & DQN \cite{DQN}                    & 0.527          & 0.551          & 0.558          & 0.564          & 0.764          & 0.778          & 0.785          & 0.793          \\ \hline
\multirow{3}{*}{Deep Binary Hashing}  & SUBIC \cite{SUBIC}                  & 0.635          & 0.689          & 0.713          & 0.721          & 0.652          & 0.783          & 0.792          & 0.796          \\ \cline{2-10} 
                                      & NINH \cite{NINH}                    & 0.600          & 0.667          & 0.689          & 0.702          & 0.597          & 0.627          & 0.647          & 0.651          \\ \cline{2-10} 
                                      & CNNH \cite{CNNH}                   & 0.496          & 0.580          & 0.582          & 0.583          & 0.536          & 0.522          & 0.533          & 0.531          \\ \hline
\multirow{3}{*}{Product Quantization} & LOQP \cite{LOPQ}                    & 0.279          & 0.324         & 0.366          & 0.370          & 0.436          & 0.452          & 0.463          & 0.468          \\ \cline{2-10} 
                                      & OPQ \cite{OPQ}                    & 0.265          & 0.315          & 0.323          & 0.345          & 0.429          & 0.433          & 0.450          & 0.458          \\ \cline{2-10} 
                                      & PQ \cite{PQ}                     & 0.237          & 0.265          & 0.268          & 0.266          & 0.398          & 0.406          & 0.413          & 0.422          \\ \hline
\multirow{3}{*}{Binary Hashing}       & SDH \cite{SDH}                    & 0.255          & 0.330          & 0.344          & 0.360          & 0.414          & 0.465          & 0.451          & 0.454          \\ \cline{2-10} 
                                      & ITQ \cite{ITQ}                    & 0.158          & 0.163          & 0.168          & 0.169          & 0.428          & 0.430          & 0.432          & 0.435          \\ \cline{2-10} 
                                      & SH \cite{SH}                     & 0.124          & 0.125          & 0.125          & 0.126          & 0.390          & 0.394          & 0.393          & 0.396          \\ \hline
\end{tabular}
\end{adjustbox}
\caption{The mean Average Precision (mAP) scores of different hashing algorithms on experimental protocol 1.}
\label{table:Table2}
\end{table*}

\subsection{Results and Analysis}

\begin{figure}[!t]
\centering
\includegraphics[width=0.8\linewidth]{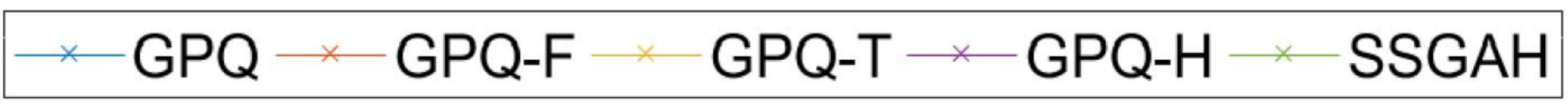}
\subfigure[CIFAR-10]{
\includegraphics[width=0.472\linewidth]{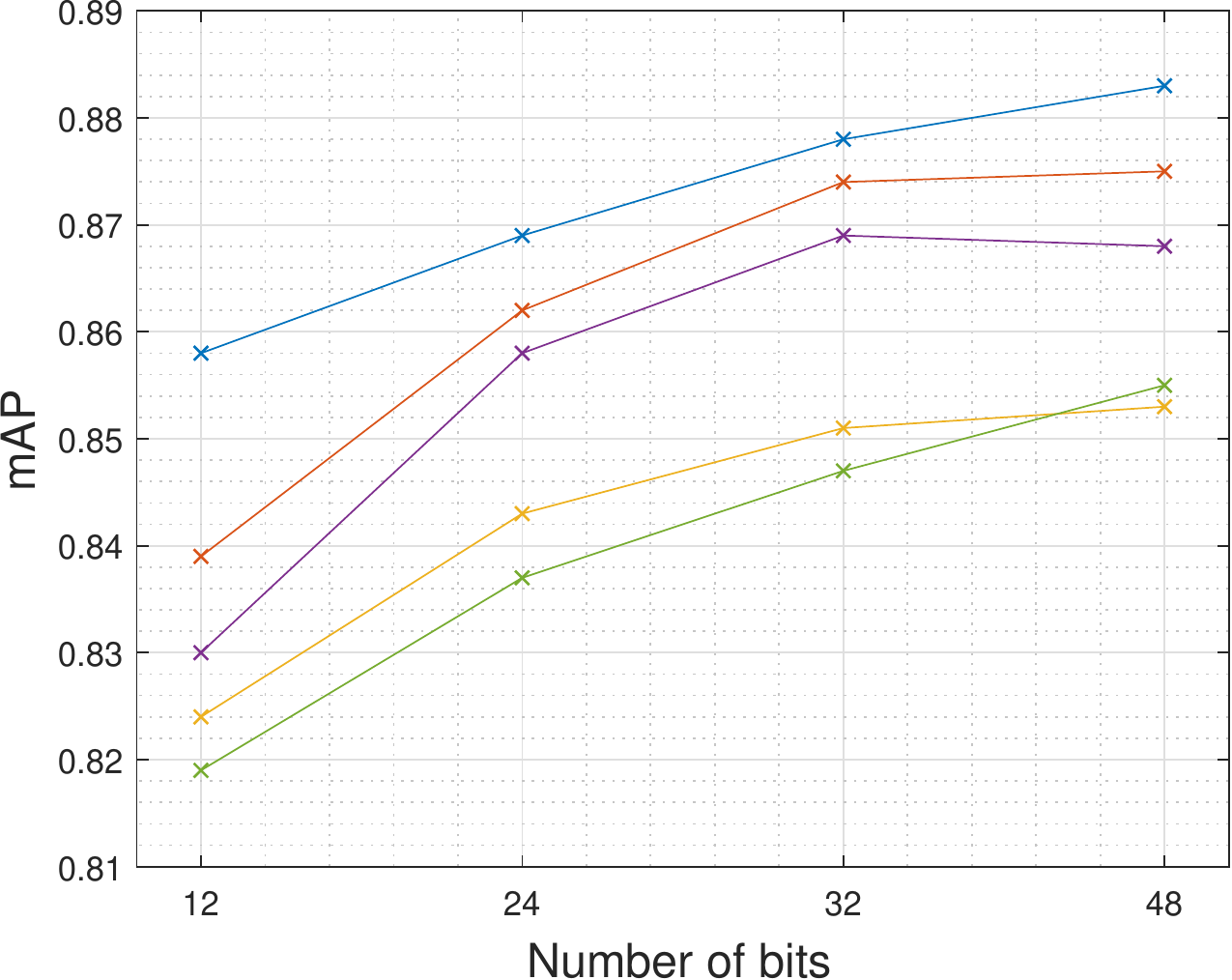}
}
\subfigure[NUS-WIDE]{
\includegraphics[width=0.472\linewidth]{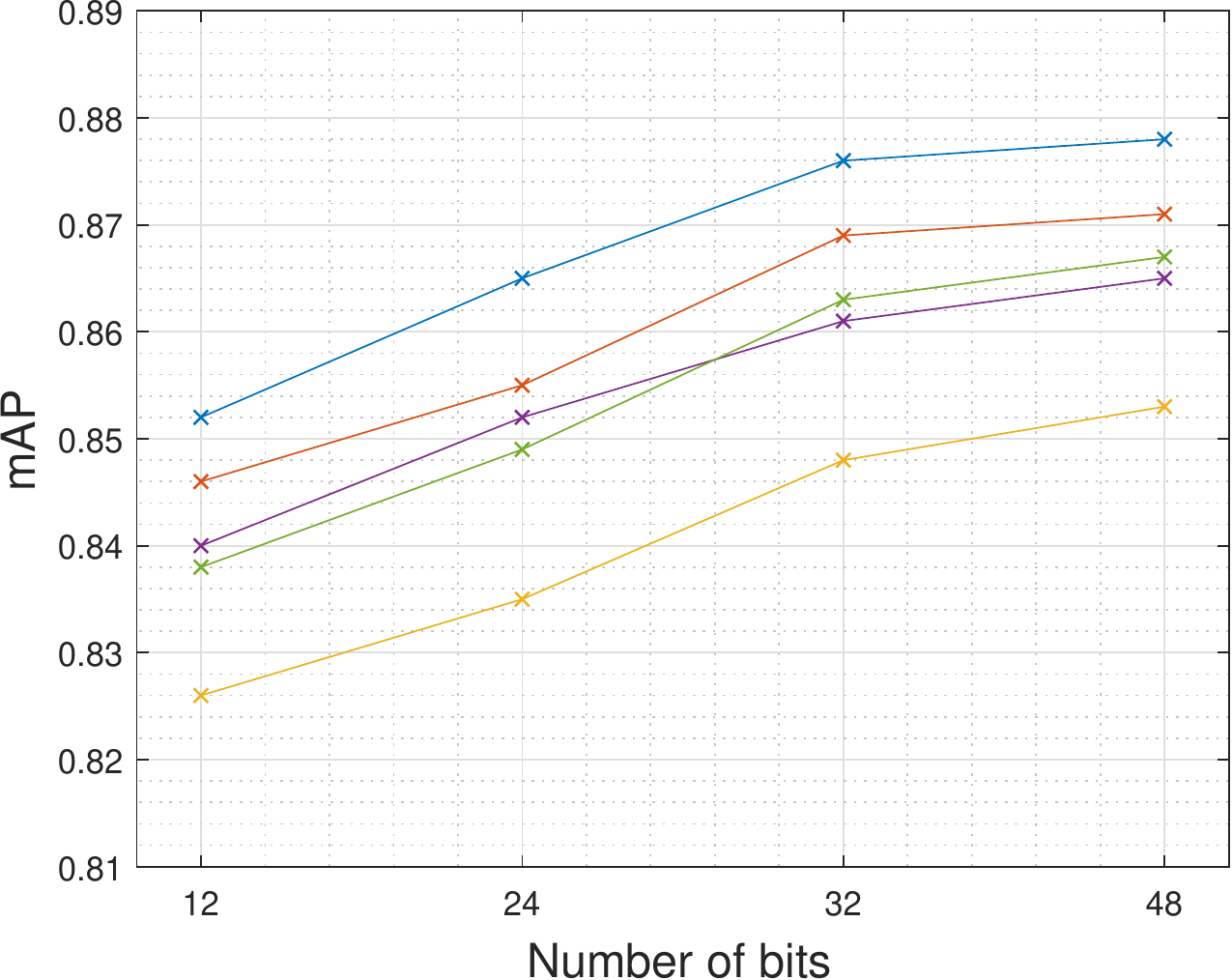}
}
\caption{The comparison results of GPQ and its variants.} 
\label{fig:long}
\label{fig:onecol}
\label{fig:Figure3}
\end{figure}

\begin{figure}[t]
\centering
\subfigure[12-bits]{
\includegraphics[width=0.472\linewidth]{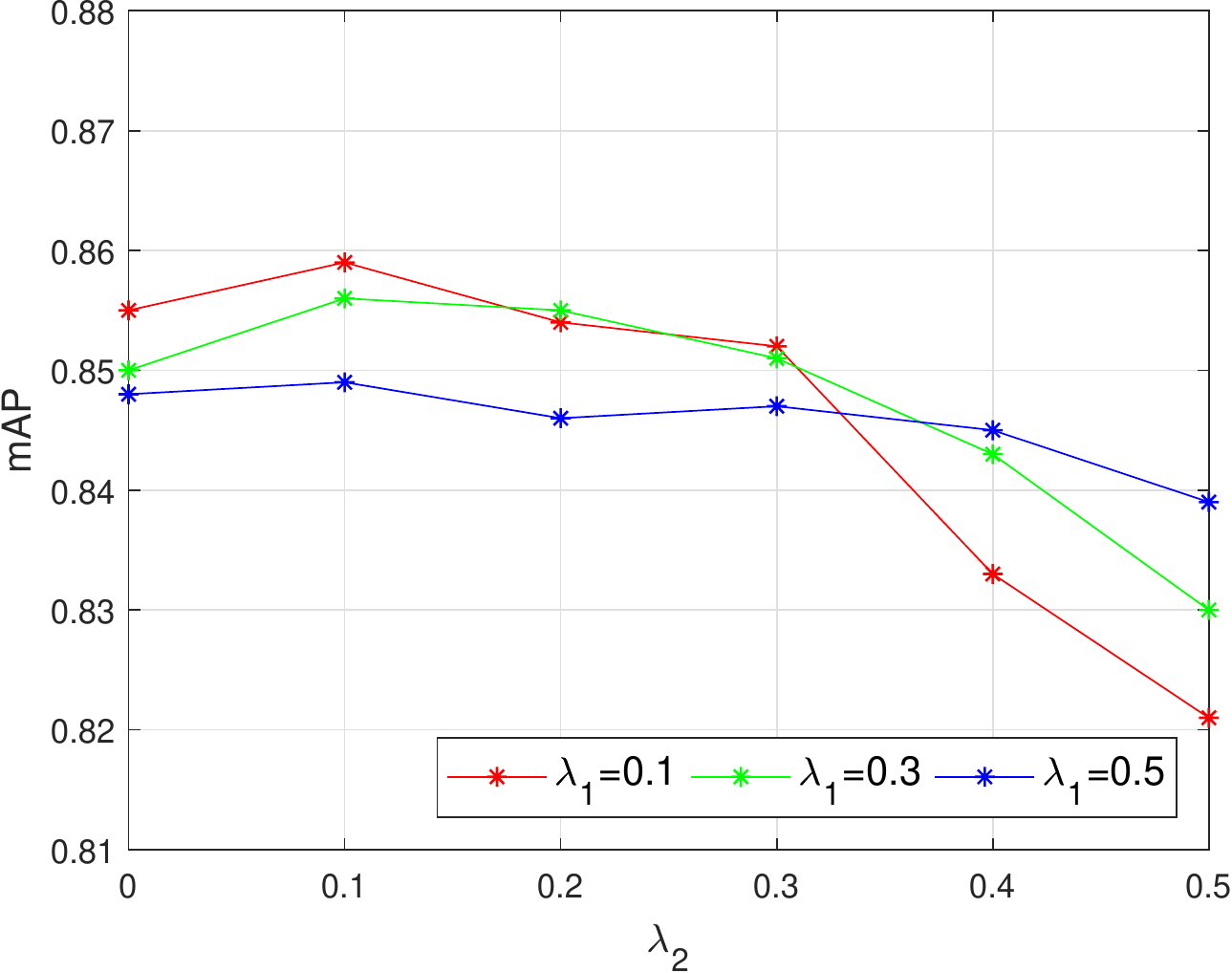}
}
\subfigure[48-bits]{
\includegraphics[width=0.472\linewidth]{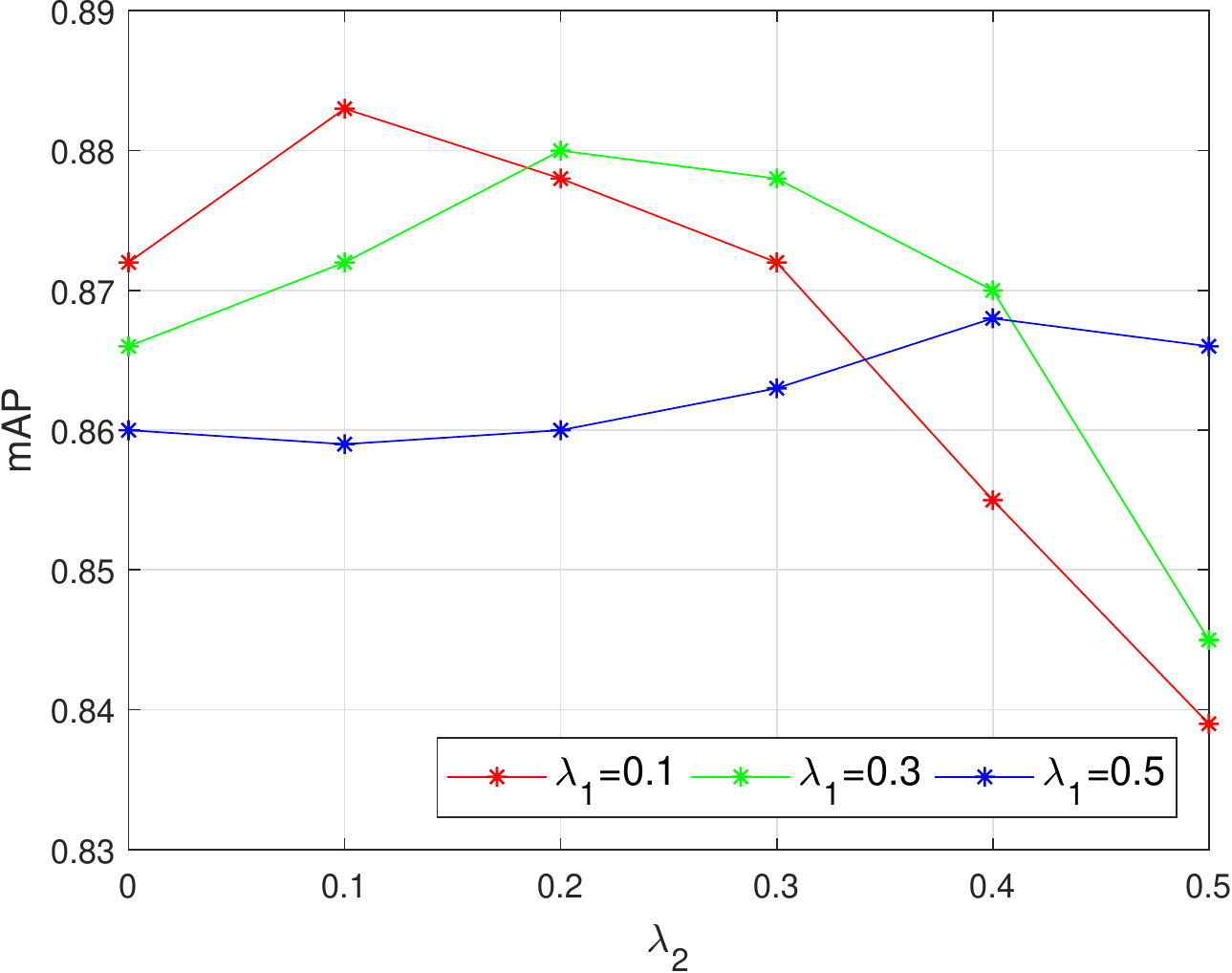}
}
\caption{The sensitivity investigation of two balancing parameters: $\lambda_1$ and $\lambda_2$.} 
\label{fig:long}
\label{fig:onecol}
\label{fig:Figure4}
\end{figure}

\begin{figure*}[!t]
\centering
\subfigure[GPQ-T]{
\includegraphics[width=0.32\linewidth]{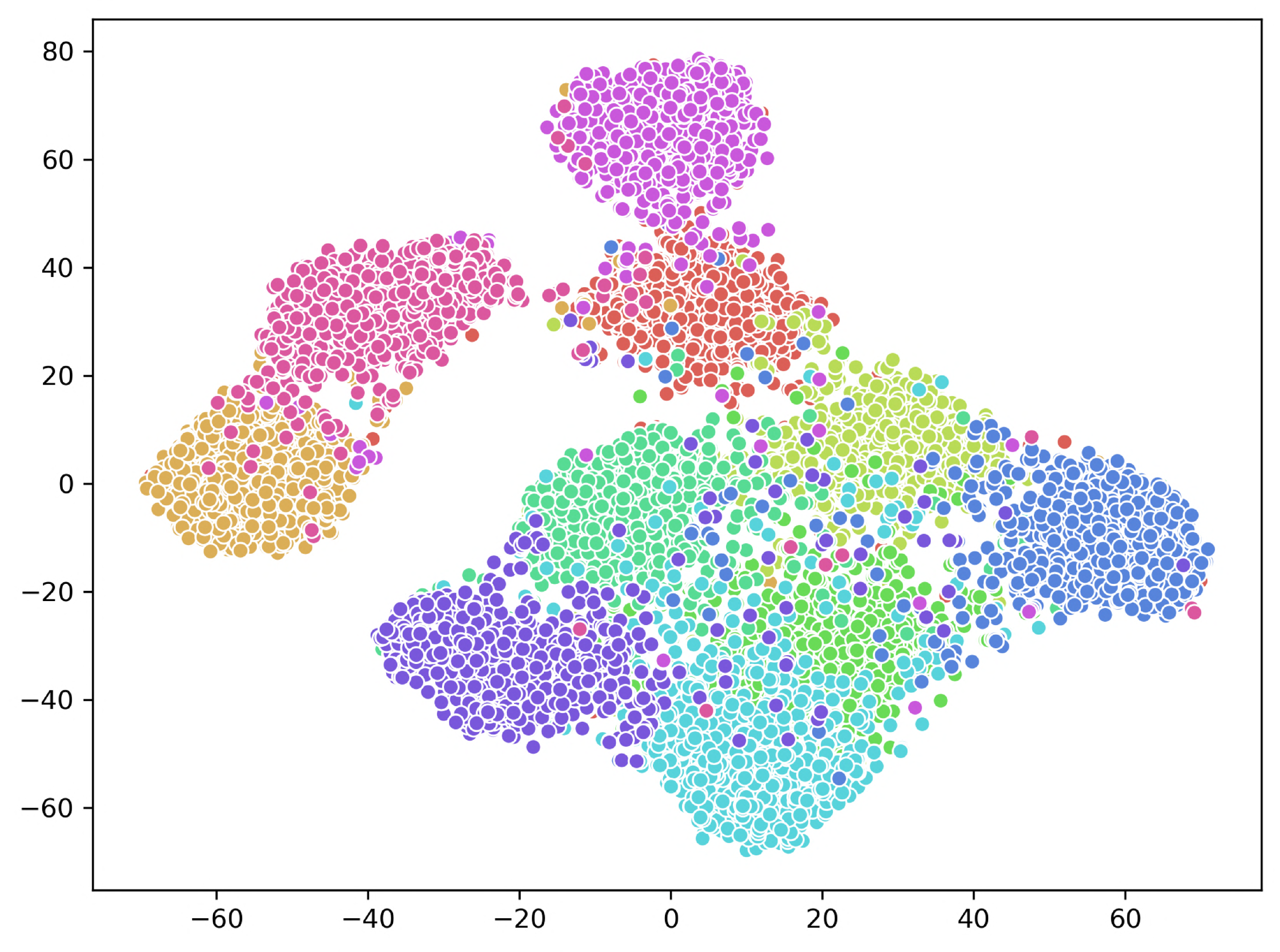}
}
\subfigure[GPQ-H]{
\includegraphics[width=0.32\linewidth]{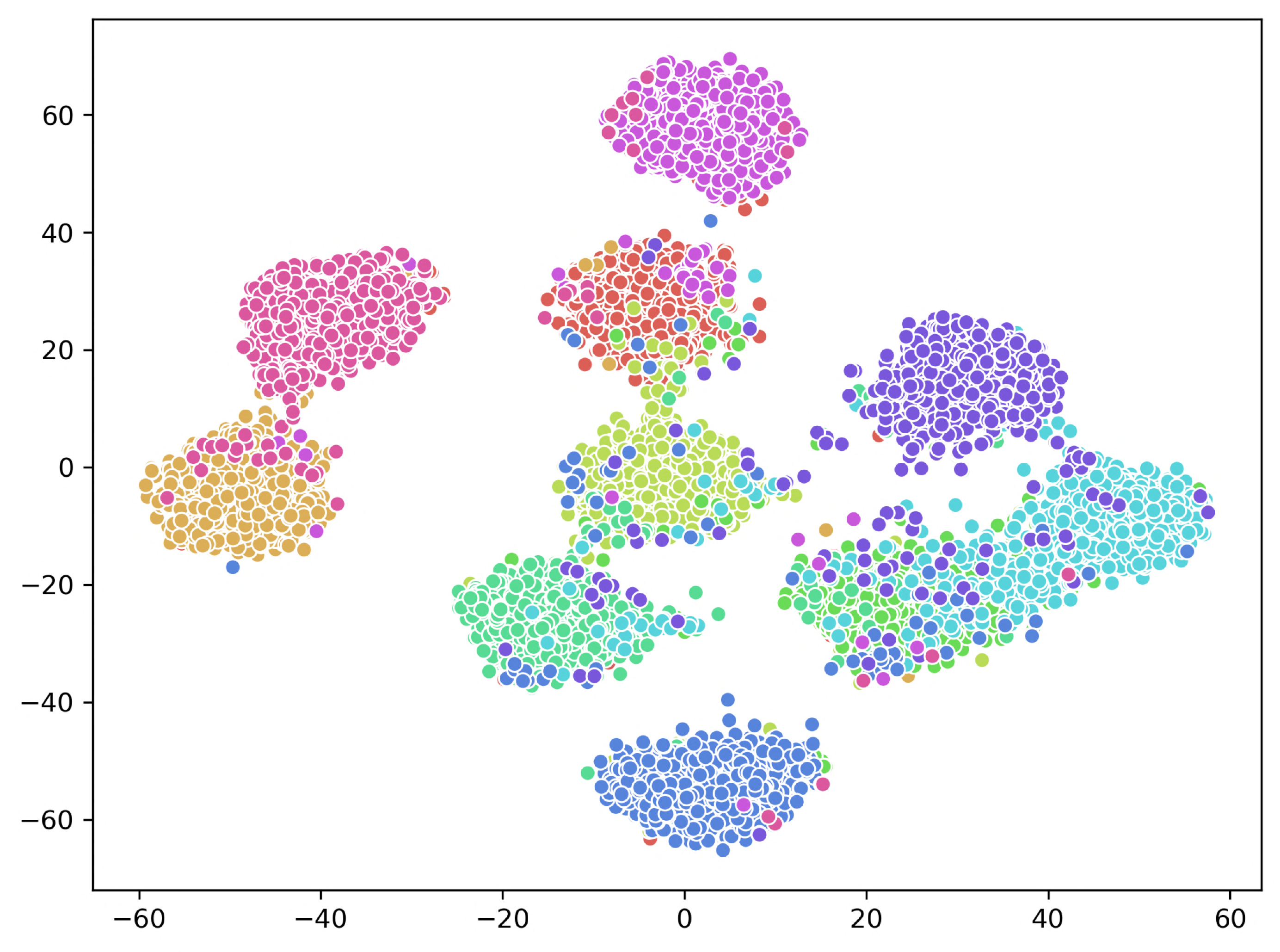}
}
\subfigure[GPQ]{
\includegraphics[width=0.32\linewidth]{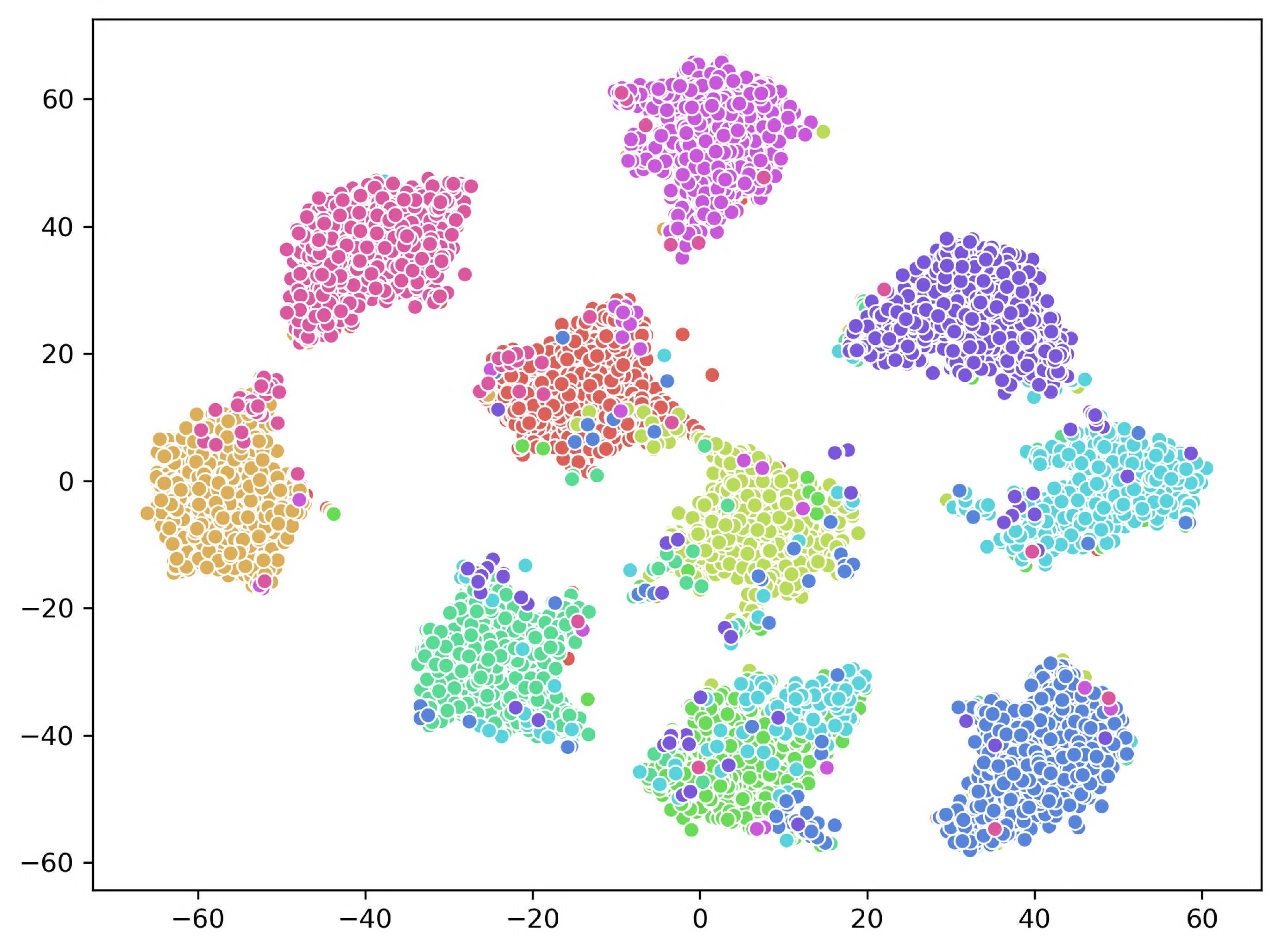}
}
\caption{The t-SNE visualization of deep representations learned by GPQ-T, GPQ-H and GPQ on CIFAR-10 dataset respectively.} 
\label{fig:long}
\label{fig:onecol}
\label{fig:Figure5}
\end{figure*}

\noindent\textbf{Overview} Experimental results for protocol 1 and protocol 2 are shown in Tables ~\ref{table:Table2} and ~\ref{table:Table3}, respectively. In each Table, the methods are divided into several basic concepts and listed by group. We investigate the variants of GPQ for ablation study, and the results can be seen in Figures ~\ref{fig:Figure3} to ~\ref{fig:Figure5}. From the results, we can observe that our GPQ scheme outperforms other hashing methods, demonstrating that the proposed loss functions effectively improve the GPQ network by training it in a semi-supervised fashion.

\bigskip

\noindent\textbf{Comparison with Others} As shown in Table ~\ref{table:Table2}, the proposed GPQ performs substantially better over all bit lengths than compared methods. Specifically, when we averaged mAP scores for all bit lengths, GPQ are 4.8\%p and 4.6\%p higher than the previous semi-supervised retrieval methods on CIFAR-10 and NUS-WIDE respectively. In particular, the performance gap is more pronounced as the number of bits decreases. This tendency is intimately related to the baseline hashing concepts. Comparing the results of the PQ-based and BH-based methods, we can identify that PQ-based ones are generally superior for both deep and non-deep cases especially for smaller bits. This is because unlike BH, PQ-based methods have the codewords of real values which enable mitigating the deviations generated during the encoding time, and they also allow more diverse distances through asymmetric calculation between database and query inputs. For the same reason, PQ-based GPQ with these advantages is able to achieve the state-of-the-art results in semi-supervised image retrieval.

\bigskip
\noindent\textbf{Ablation Study}
\label{Ablation Study}
To evaluate the contribution and importance of each component and training objectives in GPQ, we build three variants: 1) replace the feature extractor $F$ with CNN-F ~\cite{CNN-F} for \textit{GPQ-F}; 2) remove the classifier $C$ and learn the network only with the PQ table $Z$ for \textit{GPQ-H}; 3) exchange the N-pair Product Quantization loss $\mathcal{L}_{N\text{-}PQ}$ with a standard Triplet loss ~\cite{Metric2} for \textit{GPQ-T}. We perform retrieval experiments on protocol 1 for these variants, and empirically determine the sensitivity of GPQ .

The mAP scores of these variants and the original GPQ are reported in Figure ~\ref{fig:Figure3}. In this experiment, the hyper-parameters are all set identically as defaults in section ~\ref{Implementation Details}, and the balancing parameter $\lambda_1$ and $\lambda_2$ are all set to 0.1. GPQ-F employs all the training objectives and shows the best mAP scores. It even outperforms other semi-supervised retrieval methods including SSGAH ~\cite{SSGAH} in all experimental settings of protocol 1.

However, the modified VGG-based \cite{DCBH} original GPQ is superior to GPQ-F on both datasets. It reverses the general idea that CNN-F with more trainable parameters would have higher generalization capacity. This happens because high complexity does not always guarantee performance gains, same as the observations in ~\cite{Generalization}. Therefore, instead of increasing the network complexity, we focus on network structures that can extract more general features, and it results in high retrieval accuracy especially for semi-supervised situation.

To determine the contributions of Classification loss $\mathcal{L}_{cls}$ and Subspace Entropy Mini-max loss $\mathcal{L}_{SEM}$, we carry out experiments with varying the balancing parameters of each loss function $\lambda_1$ and $\lambda_2$, respectively. GPQ-H is equivalent to both $\lambda_1$ and $\lambda_2$ being zero, and the result is reported in Figure ~\ref{fig:Figure3}. Experimental results for the different options for $\lambda_1$ and $\lambda_2$ are detailed in Figure ~\ref{fig:Figure4} for the bit lengths of 12 and 48. In general, high accuracy is achieved when the influences of $\mathcal{L}_{cls}$ and $\mathcal{L}_{SEM}$ are similar, and optimal results are obtained when both of two balancing parameters are set to 0.1 for 12 and 48 bit lengths.

\begin{table*}[!t]
\centering
\begin{adjustbox}{width=0.9\textwidth}
\tiny
\begin{tabular}{|c|c|c|c|c|c|c|c|c|c|}
\hline
\multirow{2}{*}{Concept}                                                                   & \multirow{2}{*}{Method} & \multicolumn{4}{c|}{CIFAR-10}         & \multicolumn{4}{c|}{NUS-WIDE}         \\ \cline{3-10} 
                                                                                           &                         & 12-bits & 24-bits & 32-bits & 48-bits & 12-bits & 24-bits & 32-bits & 48-bits \\ \hline
\multirow{5}{*}{Deep Hashing}                                                              & GPQ (Ours)                     & \textbf{0.321}   & \textbf{0.333}   & \textbf{0.350}   & \textbf{0.358}   & \textbf{0.554}   & \textbf{0.565}   & \textbf{0.578}   & \textbf{0.586}   \\ \cline{2-10} 
                                                                                           & SSGAH \cite{SSGAH}                  & 0.309   & 0.323   & 0.341   & 0.339   & 0.539   & 0.553   & 0.565   & 0.579   \\ \cline{2-10} 
                                                                                           & SSDH \cite{SSDH}                   & 0.285   & 0.291   & 0.311   & 0.325   & 0.510   & 0.533   & 0.549   & 0.551   \\ \cline{2-10} 
                                                                                           & NINH \cite{NINH}                   & 0.241   & 0.249   & 0.253   & 0.272   & 0.484   & 0.483   & 0.485   & 0.487   \\ \cline{2-10} 
                                                                                           & CNNH \cite{CNNH}                   & 0.210   & 0.225   & 0.227   & 0.231   & 0.445   & 0.463   & 0.471   & 0.477   \\ \hline
\multirow{4}{*}{\begin{tabular}[c]{@{}c@{}}CNN features +\\ non-Deep Hashing\end{tabular}} & SDH \cite{SDH}                     & 0.185   & 0.193   & 0.199   & 0.213   & 0.471   & 0.490   & 0.489   & 0.507   \\ \cline{2-10} 
                                                                                           & ITQ \cite{ITQ}                    & 0.157   & 0.165   & 0.189   & 0.201   & 0.488   & 0.493   & 0.508   & 0.503   \\ \cline{2-10} 
                                                                                           & LOPQ \cite{LOPQ}                   & 0.134   & 0.127   & 0.126   & 0.124   & 0.416   & 0.386   & 0.380   & 0.379   \\ \cline{2-10} 
                                                                                           & OPQ \cite{OPQ}                    & 0.107   & 0.119   & 0.125   & 0.138   & 0.341   & 0.358   & 0.371   & 0.373   \\ \hline
\end{tabular}
\end{adjustbox}
\caption{The mean Average Precision scores (mAP) of different hashing algorithms on experimental protocol 2.}
\label{table:Table3}
\end{table*}

In order to investigate the effect of the proposed metric learning strategy, we substitute N-pair Product Quantization loss with triplet loss and visualize each result in Figure ~\ref{fig:Figure5}. While learning GPQ-T, the margin value is fixed at 0.1, and anchor, positive and negative pair are constructed using quantized vectors of different images. We utilize t-SNE algorithm ~\cite{t-SNE} to examine the distribution of deep representations extracted from 1,000 images for each category in CIFAR-10. From the Figure where each color represents a different category, we can observe that GPQ better separates the data points.

\noindent\textbf{Transfer Type Retrieval} Following the transfer type investigation proposed in \cite{Protocol2}, we conduct experiments under protocol 2. Datasets are divided into two partitions of equal size, each of which is assigned to train-set and test-set, respectively. To be specific with each dataset, CIFAR-10 takes 7 categories and NUS-WIDE takes 15 categories to construct train75 and test75, and the rest categories for train25 and test25 respectively. For further comparison with hashing methods of non-deep learning concept ~\cite{SDH,ITQ,LOPQ,OPQ}, we employ $7\text{-}th$ fully-connected features of pre-trained AlexNet as inputs and conduct image retrieval.

As we can observe in table ~\ref{table:Table3}, the average mAP score decreased compared to the results in protocol 1, because the label information of unseen categories disappears. In consequence, there is a noticeable mAP drop in supervised-based schemes, and the performance gap between the supervised and unsupervised learning concepts decreased. However, our GPQ method still outperforms other hashing methods. This is because, in GPQ, labeled information of known categories is fully exploited to learn the discriminative and robust codewords via metric learning algorithm, and at the same time, unlabeled data is fully explored to generalize the overall architecture through entropy control of feature distribution.

\section {Conclusion}

In this paper, we have proposed the first quantization based deep semi-supervised image retrieval technique, named Generalized Product Quantization (GPQ) network. We employed a metric learning strategy that preserves semantic similarity within the labeled data for the discriminative codebook learning. Further, we compute an entropy for each subspace and simultaneously maximize and minimize it to embed underlying information of the unlabeled data for the codebook regularization. Comprehensive experimental results justify that the GPQ yields state-of-the-art performance on large-scale image retrieval benchmark datasets.

\paragraph{Acknowledgement}

This work was supported in part by Institute of Information $\&$ communications Technology Planning $\&$ Evaluation (IITP) grant funded by the Korea government (MSIT) (No. 1711075689, Decentralised cloud technologies for edge$\/$IoT integration in support of AI applications), and in part by Samsung Electronics Co., Ltd.


\newpage

{\small
\bibliographystyle{ieee_fullname}
\bibliography{egbib}
}

\end{document}